\documentclass[12pt,letterpaper]{article}
\usepackage[a4paper, total={7in, 10in}]{geometry}

\usepackage{helvet}
\usepackage{authblk}
\usepackage{hyperref}
\usepackage{amsmath}
\usepackage{amssymb}
\usepackage{graphicx}

\makeatletter
\renewcommand{\maketitle}{\bgroup\setlength{\parindent}{0pt}
\begin{flushleft}
  \textbf{\@title}
    \\ \vspace*{1cm} 
  \@author
\end{flushleft}\egroup}
\makeatother

\newcommand{\thetab}{{\boldsymbol{\theta}}}

\newcommand{\s}{\mathbf{s}}
\newcommand{\x}{\mathbf{x}}
\newcommand{\y}{\mathbf{y}}
\newcommand{\n}{\mathbf{n}}

\newcommand{\f}{\mathbf{f}}
\newcommand{\g}{\mathbf{g}}

\newcommand{\J}{\mathbf{J}}

\newtheorem{theorem}{Theorem}

\renewcommand{\v}{{\bf v}}
\newcommand{\xb}{{\bf x}}
\newcommand{\ub}{{\bf u}}
\newcommand{\zb}{{\bf z}}

\newcommand{\hb}{{\bf h}}
\newcommand{\defeq}{:=}
\newcommand{\dd}{\text{d}}

\title{\Large Nonlinear Independent Component Analysis for Principled Disentanglement in Unsupervised Deep Learning}
\date{}

\author[1*]{Aapo Hyv\"arinen}
\author[2]{Ilyes Khemakhem}
\author[3]{Hiroshi Morioka}

\affil[1]{Department of Computer Science, University of Helsinki}
\affil[2]{Gatsby Computational Neuroscience Unit, University College London}
\affil[3]{RIKEN AIP}
\affil[*]{Correspondence: aapo.hyvarinen@helsinki.fi}

\usepackage[super,comma,sort&compress]{natbib}

\begin{document}

\maketitle

\section*{Summary} 

A central problem in unsupervised deep learning is how to find useful representations of high-dimensional data, sometimes called "disentanglement". Most approaches are heuristic and lack a proper theoretical foundation. In linear representation learning, independent component analysis (ICA) has been successful in many applications areas, and it is  principled, i.e., based on a well-defined probabilistic model. However, extension of ICA to the nonlinear case has been problematic due to the lack of identifiability, i.e., uniqueness of the representation. Recently, nonlinear extensions that utilize temporal structure or some auxiliary information have been proposed. Such models are in fact identifiable, and consequently, an increasing number of algorithms have been developed. In particular, some self-supervised algorithms can be shown to estimate nonlinear ICA, even though they have initially been proposed from heuristic perspectives. This paper reviews the state-of-the-art of nonlinear ICA theory and algorithms.

\section*{Keywords}

Unsupervised learning ; representation learning ; disentanglement ; independent component analysis ; nonlinear ICA

\section*{Introduction}

Recent advances in data collection have resulted in very large data sets,
including images
\citep{lecun1998gradientbased,krizhevsky2012imageneta,deng2009imagenet},
3D shapes~\citep{chang2015shapenet},
text~\citep{marcus1993building,maas2011learning},
music~\citep{bertin-mahieux2011million},
and graphs and networks~\citep{hu2020open,yanardag2015deep}.
As the amount and complexity of the
data started growing, most of the work in machine learning research went towards
developing preprocessing pipelines to assist the extraction of meaningful
information from large data sets, allowing for efficient learning.  With the
rise of deep learning, preprocessing shifted from hand-crafted, expertise-based
feature engineering to utilizing neural networks to implicitly learn useful
representations.  This is known as \textit{representation learning}, and it
has grown to be one of the pillars of modern machine learning.
Representations learned by deep neural networks are now
widely used in many machine learning applications, including speech recognition
and processing~\citep{dahl2011contextdependent,seide2011conversational},
natural language processing~\citep{bengio2003neural,devlin2019bert}, %
action recognition~\citep{korbar2018cooperative}, domain adaptation~\citep{wang2018deep}, and many more.

Learning good representations can have a significant impact on the performance
of subsequent machine learning~\citep{bengio2013representation}.
Representation learning can sometimes be based on supervised learning with labelled data, in which case the representation is transferred to a new data set. But since  labelling is a costly and time-consuming endeavor and only a small percentage of today's data sets are labelled, it would be better to learn the representation without any labels or targets, that is, in an unsupervised way.

The quality of a learned representation is frequently characterized by its capacity to
improve the performance of a ``downstream'' task in which the user is currently
engaged. This criterion, however, is only meaningful when such a task exists
and is clearly defined; typically, it consists of classification or regression
on a labelled data set. 
However, different representations may be optimal for different classification tasks.
It would be better to be able to assess the quality of a representation by a criterion that is inherent to the representation itself, rather than reliant on the context or task in which
it may be employed. Here, we consider the problem of finding a generally useful representation based on unsupervised learning.

Well-known unsupervised methods, including variational autoencoders (VAE)~\citep{kingma2014autoencoding,rezende2014stochastic} and normalizing flows~\citep{kobyzev2020normalizing}, learn a posterior
distribution over a possibly
lower-dimensional latent variable.  It is hoped that such a posterior will
correspond to the underlying distribution of statistically independent sources
of variation.
A related line of research is being developed for the related goal of learning \textit{disentangled representations}~\citep{higgins2017betavae,alemi2017fixing,burgess2018understanding,chen2018isolating,esmaeili2019structured,mathieu2018disentangling,kim2018disentangling}. The objective is to isolate the influence of all factors of variation, which again translates to learning a representation with independent components~\citep{bengio2013representation,burgess2018understanding}. Many methods thus learn disentangled representations by imposing independence on the latent variables and adding regularization terms to the VAE objective in an ad-hoc manner~\citep{zhao2017infovae,gao2019autoencoding,achille2018information,kumar2017variational,esmaeili2019structured}.

A recent line of research aims to go further than mere independence by  learning representations
that are true to the explanatory factors of variation behind the data.
This desideratum is formalized by the notion of \textit{identifiability}.
Fundamentally, an identifiable %
probabilistic model
can only learn one representation in the limit of infinite data: the ground
truth generative factors.
Identifiability is thus necessary for learning representations that are
semantically meaningful, reproducible, interpretable and better suited for
downstream tasks~\citep{bengio2013representation,peters2017elements,schmidhuber1996semilinear}.

Unfortunately, the above-mentioned techniques do not allow for any
theoretical identifiability guarantees.
In fact, disentangled representations are not identifiable in general.
In other words, learning nonlinear models that seek independence results in
arbitrary representations that are not always related to the ground truth
factors of variation.
A large scale empirical study~\citep{locatello2019challenging} showed that the proposed models for disentanglement exhibit substantial variance depending on hyperparameters and random seeds.
Unsupervised learning of identifiable nonlinear
representations has long been known to be theoretically impossible~\citep{Hyva99NN,locatello2019challenging} without any ``inductive biases'', i.e., suitable constraints on the model.

Within representation learning, identifiability has mostly been studied in the
context of \textit{independent component analysis} (ICA).
In ICA, the observations are considered to be a mixture of independent latent components.  The goal is to learn an ``demixing'' transformation capable of recovering the original components based on their independence and the observed mixed data. In the linear case, the theory and algorithms are already quite developed~\citep{Hyva00NN,Hyvabook}, while nonlinear versions of ICA are quite recent. The promise is that being probabilistic and identifiable, nonlinear ICA is a general, principled solution for the problem of disentanglement.

Meanwhile, recent work in computer vision has successfully proposed ``self-supervised'' feature extraction methods from a purely heuristic perspective. The idea is to reformulate the unsupervised learning problem as a supervised learning problem using a judiciously defined ``pretext'' task.
One fundamental example is to train a neural network to discriminate the observed, unlabelled data from some artificially generated noise~\citep{Gutmann12JMLR}.
A large number of heuristic methods have further been proposed based on the intuitively comprehensible structure of images ~\citep{misra2016shuffle,noroozi2016unsupervised,larsson2017colorization}.  
As such, self-supervised learning (SSL) has the potential of providing computationally efficient algorithms for disentanglement. Empirically, such approaches have allowed unsupervised learning to be leveraged for supervised tasks resulting in dramatic performance improvements. However,  it is widely  acknowledged that most such methods lack theoretical grounding. Since such methods are not necessarily based on probabilistic modelling, the question of identifiability cannot always be meaningfully approached, although uniqueness can be considered from a more general perspective~\citep{d2020underspecification}. Ideally, we would like to combine SSL with probabilistic modelling and achieve identifiability.

In this paper, we review recent methods for unsupervised representation
learning that aim to learn the ground truth generative factors. Thus, we focus on probabilistic models which are identifiable, the main framework being nonlinear ICA.
Particular emphasis will here be put on algorithms, especially of the self-supervised kind. (See our companion paper~\citep{Hyva23AISM} for a more theoretical treatment of identifiability.)


\section*{Background: Linear ICA}
\label{sec:intro:lin_ica}

Over the decades, ICA has been
extensively studied in the linear setting, where the mixing is considered to be
a performed by a matrix~\citep{comon1994independent,Hyva00NN,Hyvabook,Cardoso01}. %
Linear ICA has applications in neuroscience, including functional magnetic resonance imaging (fMRI)~\citep{mckeown1998analysis,calhoun2003ica,Beckmann05}
and EEG/MEG~\citep{delorme2007enhanced,milne2009independent,Brookes11, Hyva10NI},
document analysis~\citep{bingham2002ica,podosinnikova2015rethinking}, finance~\citep{Back,oja2000independent}, astronomy~\citep{nuzillard2000blind}, image processing~\citep{Hyvanisbook}
and many more fields.  The central theoretical result is that if all the latent components (``sources'') are
\textit{non-Gaussian}, linear ICA is identifiable.

Consider a vector of latent variables $\mathbf{s}=(s_1,\ldots,s_d)$ which is transformed through an unknown linear mixing into observations $\mathbf{x}$:
\begin{equation}
        \label{eq:intro:ica}
        \mathbf{x} = \mathbf{A} \mathbf{s},
\end{equation}
where $\mathbf{A} \in \mathbb{R}^{d\times d}$ is an invertible ``mixing'' matrix.
We want to know if we can recover the original but unknown signals $s_i$ while making no or only very weak assumptions on its distribution.
Both the distribution of $\mathbf{s}$ and the mixing matrix $\mathbf{A}$ are unknown, making it difficult to determine whether a good fit to the data is related to the true generative process. This problem is also known as blind source separation (BSS).

A well-known result is that if $\s$ is Gaussian, we cannot recover it from the mixtures. This is easy to prove. It is enough to consider the special case where $\s$ is constrained to be white in the sense that the $s_i$ are uncorrelated variables with unit variance. Then, any orthogonal transformation of the components has exactly the same distribution, which is due to the rotational symmetry of the white Gaussian distribution. Its probability density function (pdf) is $p(\s)\propto \exp(\|\s\|^2/2)$ which only depends on the norm; it will not change if $\s$ is transformed by an orthogonal transformation. Thus, an arbitrary orthogonal transformation could always be made on $\s$, resulting in exactly the same observed distribution, so that orthogonal transformation cannot be determined from the data.

The framework of Independent Component Analysis (ICA)~\citep{Jutten91,comon1994independent,Hyvabook} provides a solution to this problem by making two  assumptions.
First, the components $s_1, \ldots s_d$ of the latent vector $\mathbf{s}$ are statistically \textit{independent}. This means the pdf's factorize as
       \begin{equation} \label{factorize}
                p(\mathbf{s}) = \prod_i p_i(s_i).
        \end{equation}
        Second, and most importantly, we make the assumption that all the components have \textit{non-Gaussian} distributions (except perhaps one).
Under these assumptions, the model~(\ref{eq:intro:ica}) is identifiable, meaning that the linear mixing, as well as the true components, can be estimated.
Linear ICA achieves this goal by learning an \textit{demixing} matrix $\mathbf{B}$ such that $     \mathbf{z}:=\mathbf{B}\mathbf{x}$  has statistically independent components: it can be proven that only the true components are independent and no mixtures can be, if the true components are non-Gaussian as well as independent.
We note that still, the linear ICA problem has some minor indeterminacies: The ICA model does not determine the permutation (ordering), the scaling or the signs of the independent components.

\begin{figure}
\begin{center}
  \resizebox{\textwidth}{!}{\includegraphics{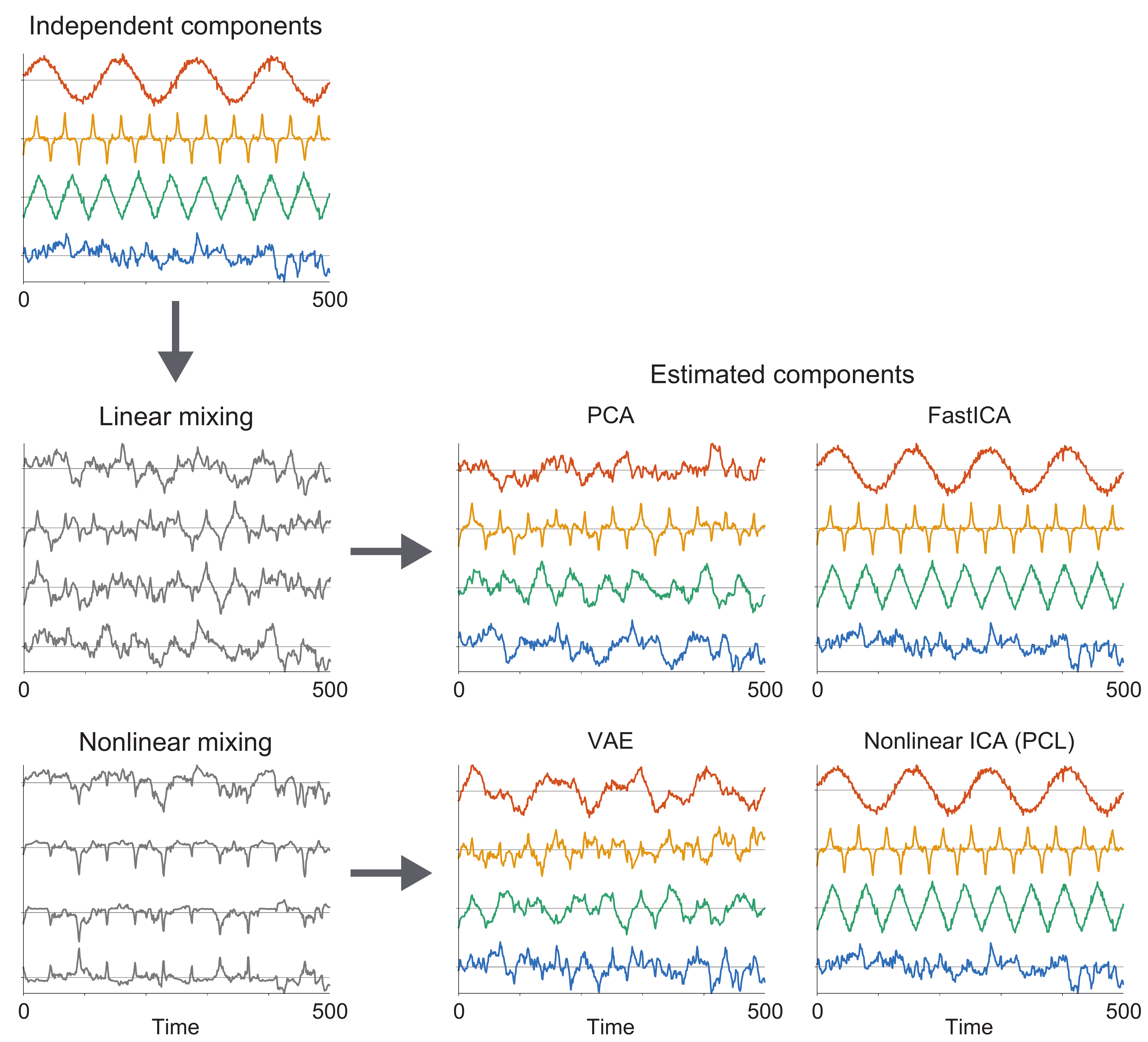}}
  \caption{Identifiability of ICA and its application on blind source separation illustrated. The original signals (top row) are mixed either linearly or nonlinearly, in the middle and bottom rows, respectively. Then linear ICA (FastICA) or nonlinear ICA (in this case, PCL) is applied on those two mixtures.  Such methods do recover the original signals, as seen in the right-most column. For comparison, PCA and its nonlinear counterpart, VAE, are applied on the same mixtures in the middle column, and we see that separation is not achieved. } \label{separation.fig}
\end{center}
\end{figure}

The idea of blind source separation is illustrated in Fig.~\ref{separation.fig}. We have four original signals which have visually nice shapes for the purpose of this illustration. They are linearly mixed, and we apply principal component analysis (PCA) and linear ICA on them (middle row of Fig.~\ref{separation.fig})). PCA does not recover the original signals, while ICA does. Next we consider the case of nonlinear mixtures, i.e., nonlinear ICA, which is already alluded to at the bottom row of that figure.

\section*{Nonlinear ICA: Problem of Identifiability}
\label{sec:intro:nica}

A straightforward generalization of ICA to the nonlinear setting would assume
that the independent components are mixed into an
observed data vector through an arbitrary but usually smooth transformation.
The matrix $\mathbf{A}$ in the linear ICA model in Eq.~(\ref{eq:intro:ica}) is replaced by an invertible mixing function
$\mathbf{f}: \mathbb{R}^d \to \mathbb{R}^d$:
\begin{align}
        \label{NICA}
                \mathbf{x} &= \mathbf{f}(\mathbf{s}).
\end{align}
The goal of nonlinear ICA is to learn an demixing function $\mathbf{g}$ that generalizes
the demixing matrix $\mathbf{B}$ such that
\begin{equation}
        \label{NICA_g}
\mathbf{z}\defeq\mathbf{g}(\mathbf{x})
\end{equation}
gives the original independent components as $\zb=\s$. Such a nonlinear mixing is illustrated in Fig.~\ref{separation.fig}, bottom row.

In the linear setting,
solving the problem of recovering the original signal $\mathbf{s}$ is equivalent to finding statistically independent components as we saw above.
However, a fundamental problem with nonlinear ICA is that solutions to Eq.~(\ref{NICA_g})
such that $\mathbf{z}$ has independent components
exist, and they are highly non-unique.
In fact, in the nonlinear case,
identifiability is a far more difficult aim to achieve. Nonlinear transformations introduce numerous degrees of freedom, rendering the problem ill-defined.

Unlike  in the linear case,
two non-Gaussian independent components $s_i$ and $s_j$ can be mixed nonlinearly while remaining statistically independent.
Equivalently, it is 
possible to explicitly construct a representation $\mathbf{z} = \mathbf{g}(\mathbf{x})$
with independent components that is
nonetheless a nonlinear mixture of the underlying independent generative
factors~\citep{Hyva99NN}.
This construction can be traced back to Darmois' work in the 1950's~\citep{darmois1953analyse},
which showed that for any two independent random variables $\xi_1, \xi_2$, we can
construct infinitely many random variables $y_1 = f_1(\xi_1,\xi_2)$ and $y_2 =
f_2(\xi_1, \xi_2)$ that are also independent.
This fundamental unidentifiability result is summarized by the following theorem:\citep{Hyva99NN}

\begin{theorem}
\label{th:intro:nica_niden}
       Let $\mathbf{x}$ be a random vector of any distribution. Then there exists a transformation
       $\mathbf{g}:\mathbb{R}^d\to[0,1]^d$
       such that
       $\mathbf{z} = \mathbf{g}(\mathbf{x})$ has a uniform distribution.
       In particular, the components $z_i \in \mathbf{z}$ are independent. Furthermore, the function $\mathbf{g}$ can be chosen so that the first variable is simply transformed by a scalar function: $z_1=g_1(x_1)$.
\end{theorem}

\begin{figure}
\begin{center}
   \includegraphics[width=0.6\columnwidth]{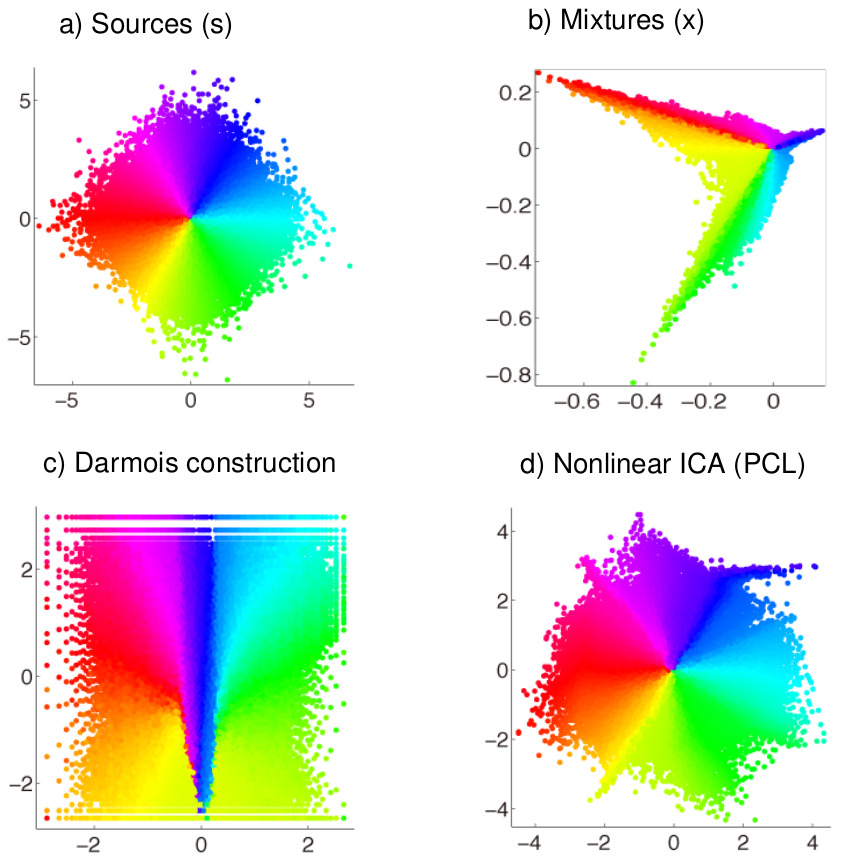}\\
\end{center}
\caption{Illustration of the unidentifiability of the basic formulation of nonlinear ICA. a) Scatterplot of two original independent components. The points are color-coded merely for the purpose of this illustration. b) A nonlinear mixing of those two independent components. c) Two estimated components obtained by the Darmois construction in Theorem 1. The components are independent, but clearly not equal to the original independent components. d) The components estimated by an identifiable version nonlinear ICA (PCL, based on temporal structure, explained later in the text); these components are a good match to the original components.}  \label{ident.fig}
\end{figure}

The function $\mathbf{g}$ in Theorem~\ref{th:intro:nica_niden} is constructed through
an iterative procedure analogous to Gram-Schmidt orthogonalization, by
recursively applying the conditional cumulative distribution function (CDF) of
$\mathbf{x}$:
\begin{equation}
        \label{eq:intro:darmois}
        z_i = g_i(x_{1},\ldots,x_i) \defeq \int_{-\infty}^{x_i} p(\tilde{x}_i \vert x_{1},\ldots,x_{i-1}) \dd \tilde{x}_i.
\end{equation}
This theorem indicates that nonlinear ICA is unidentifiable. One way is to notice that the $z_i$ can be easily point-wise transformed into independent Gaussian variables (by putting them through the inverse Gaussian cdf), and then the rotational indeterminacy holds as in the linear case~\citep{Hyva99NN}.
Another is to construct examples where it is clear that $z_i$ obtained by Equation~(\ref{eq:intro:darmois}) are not equal to the original $s_i$ even up to some nonlinear scaling indeterminacies. In particular, since $z_1 = g_1(x_1)$, as is clear from (\ref{eq:intro:darmois}), we %
would conclude that  $x_1$ is always one of the independent components,
which is absurd. The unidentifiability is illustrated in Fig.~\ref{ident.fig}~\textsf{a)-c)}.

Thus, as far as disentanglement is considered to mean finding the original components $\s$ in a nonlinear mixing such as Eq.~(\ref{NICA}), the very problem seems to be ill-defined. This is a fundamental problem which is receiving increasing attention in the deep learning community, and forms the basic motivation for nonlinear ICA theory.

 \subsection*{Variational autoencoders}
 We next point out how the problem of unidentifiability concerns  VAEs~\citep{kingma2014autoencoding,rezende2014stochastic},
 which serve as the foundation for most of the recent disentanglement methods
~\citep{higgins2017betavae,burgess2018understanding,chen2018isolating,gao2019autoencoding,esmaeili2019structured,mathieu2018disentangling,kim2018disentangling,zhao2017infovae,achille2018information}.
 We suppose that the observation $\mathbf{x}$ is generated by a latent
 vector, which we denote by $\mathbf{z}$ as usual in that context. This generative process consists of sampling from a prior
 $p_\thetab(\mathbf{z})$ and then sampling from the ``likelihood''
 $p_\thetab(\mathbf{x}\vert\mathbf{z})$, also known as a \textit{decoder}; both distributions are parameterized by $\thetab$.
 In practice, most work uses a nonlinear mixing model which formally looks very much like nonlinear ICA, but with Gaussian noise added:
 \begin{equation} \label{VAE}
   \x=\f(\zb) + \n
\end{equation}
where $\n$ is Gaussian noise of covariance $\sigma^2 \mathbf{I}$. Importantly, even the latent variables $\zb$ are Gaussian and have covariance equal to identity.
Therefore, in such a deep latent variable model, the unidentifiability is even more serious and can be more easily demonstrated. %
Since the latent vector $\zb$ is assumed to be Gaussian and white (uncorrelated variables of unit variance),
even the basic unidentifiability theory of linear ICA with Gaussian components shows that the latent variables cannot be recovered. Moreover, the Darmois theory applies as well, so the unidentifiability is even worse.
We note that exactly the same could be said about generative adversarial networks (GANs). 

Nevertheless, VAE is widely used for disentanglement, i.e., finding interesting features from the data. 
Some modifications of VAEs have also been proposed with the goal of improving disentanglement~\citep{higgins2017betavae,kim2018disentangling,chen2018isolating}, typically by adding some kind of regularization. However, there is little reason to assume that such variants would solve the fundamental problem of identifiability already pointed out by Darmois in the 1950's, and aggravated by using white Gaussian latents. It is thus questionable if VAE, or most of its variants, are well-suited for the purpose of disentanglement.
Nevertheless, since the equation in (\ref{VAE}) is almost identical to the nonlinear ICA model, it is clear that some variant must be identifiable since it can eventually coincide with nonlinear ICA (which will be made identifiable below). 

 In fact, it might be more meaningful to see VAE as a nonlinear version of PCA, which is a fundamental method for dimension reduction. Just like linear PCA, VAE can perform dimension reduction quite well, but there is no guarantee that the components obtained would be meaningful individually: only the low-dimensional manifold that they define can be considered meaningful. We note that autoencoders have been used for such dimension reduction for a very long time~\citep{HechtNielsen95}. Empirically, Fig.~\ref{separation.fig} shows in the bottom row that VAE does not find the original components, i.e., it does not separate signals.

 On the other hand, VAE is also a general-purpose method for estimating deep latent variable models, and not at all restricted to the model just mentioned. The term ``VAE'' thus has two different meanings in the literature, which is sometimes confusing.  Below, we will actually discuss estimation methods based on VAE  which estimate identifiable versions of the nonlinear ICA model. Those models can in their turn be interpreted as  identifiable versions of the ``VAE model'' in (\ref{VAE}).

\section*{Nonlinear ICA: Identifiable models and algorithms}

While the results above are negative, the main point in this review is to discuss how it is in fact possible to make nonlinear ICA models identifiable. The key is to provide some additional information to the model. The Darmois construction assumes that the data points are all obtained independently of each other and have identical distributions (called ``i.i.d.\ sampling''). However, this is often not the case in reality. A fundamental case is time series, where the time points are not independent of each other since there can be, for example, autocorrelations; nor are the time points necessarily identically distributed since the time series can be nonstationary.

\begin{figure}
\begin{center}
  \resizebox{\textwidth}{!}{\includegraphics{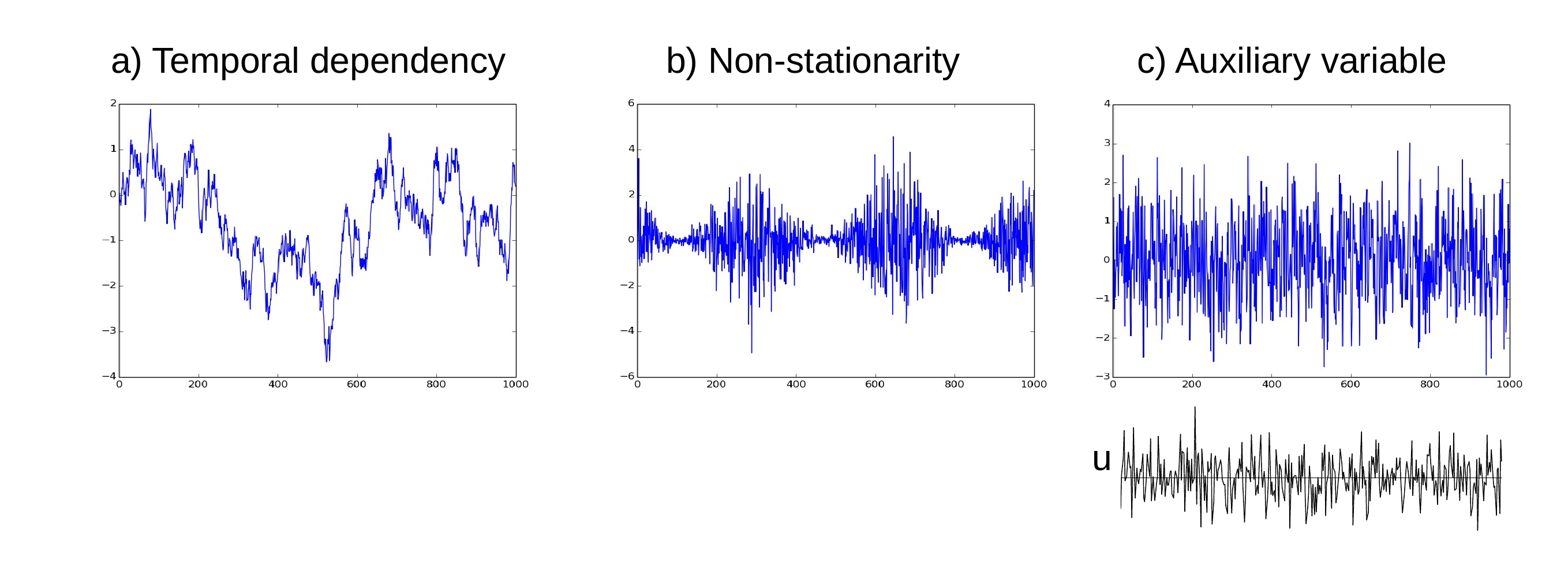}}
  \caption{Three properties of independent components that allow identifiability in nonlinear ICA: a) temporal correlations or other dependencies, b) nonstationarity, here depicted as nonstationarity of variance, c) an auxiliary variable $u$ that modulates the distribution of the component, without any temporal structure.} \label{three.fig}
\end{center}
\end{figure}

In recent years, a number of identifiability results
have been based on the temporal structure of the observed data, or, equivalently, the temporal structure of the independent components. Thus, we modify the mixing equation in (\ref{NICA}) to explicitly include the time index $t$:
\begin{align}        \label{NICA_t}
                \mathbf{x}(t) &= \mathbf{f}(\mathbf{s}(t)) 
\end{align}
Initial work assumed that the independent components are
autocorrelated time series~\citep{Harmeling03,Sprekeler14}.
Further models were subsequently proposed assuming that
the data come from nonstationary time series~\citep{Hyva16NIPS} or 
have general non-Gaussian temporal dependencies~\citep{Hyva17AISTATS}. %
However, temporal structure is not the only approach that leads to identifiability.
Alternatively, it can be that we have access to an auxiliary
variable that modulates the distributions of the independent components~\citep{Hyva19AISTATS,Khemakhem20iVAE} and leads to identifiability. The three properties  leading to identifiability are illustrated in Fig.~\ref{three.fig}. These models
achieved significant progress towards providing identifiability guarantees
by integrating side information into the generative model.
In the following, we go through the main models and learning methods.

\subsection*{Time-Contrastive Learning}
Time-Contrastive Learning (TCL)~\citep{Hyva16NIPS} is a method for nonlinear ICA based on the
assumption that while the sources are independent, they are also \textit{nonstationary}
time series. 
This implies that they can be divided into non-overlapping segments, such
that their distributions vary across segments. Such an idea is well-known in the theory of linear blind source separation~\citep{Matsuoka95,Pham01,Cardoso01}. 
The nonstationarity is supposed to be slow compared to the sampling rate,
allowing us to consider the distributions inside each segment to be constant
over time, and resulting in a piece-wise stationary process. We can give an intuitive justification for why such a model is identifiable:   We impose the estimated components to be \textit{independent at every segment}, which means we get many more independence constraints in finding the independent components. Thus, it is intuitively plausible that we get a unique solution.

Formally, given a segment index $\tau \in \{1,\ldots,T\} $ where $T$ is
the number of segments, the distribution of each latent component $s_i$ within
that segment is modelled as an exponential family:
\begin{equation}
\label{eq:intro:tcl}
\log p_\tau(s_i) = \log q_{i,0}(s_i) + \sum_{j=1}^k\lambda_{i,j}(\tau) q_{i,j}(s_i) - \log Z_i(\lambda_{i,1}(\tau), \ldots, \lambda_{i,k}(\tau)),
\end{equation}
where $q_{i,0}$ is a stationary base density and $\mathbf{q}_i \defeq (q_{i,1},\ldots,q_{i,k})$
are the sufficient statistics
for the exponential family of the component $s_i$, and $Z_i$ is the normalization constant.
Importantly, the parameters $\lambda_{i,j}(\tau)$
depend on the segment index $\tau$, indicating that the
distributions of the components change across segments. 

TCL recovers the inverse transformation $\g=\mathbf{f}^{-1}$ by \textit{self-supervised
learning}, where the pretext task is to classify original data points with segment
indices giving the labels, using multinomial logistic regression. To this end, TCL employs a deep neural network consisting of a feature extractor $\mathbf{h}(\xb; \thetab)$, with $\thetab$  parametrizing the
neural network, followed by a final classifying layer
(e.g., softmax).
Intuitively, this
is premised on the fact that in
order to optimally classify observations $\x(t)$ into their corresponding
segments $\tau$, the feature extractor $\mathbf{h}(\xb; \thetab)$
must learn about the temporal changes in the underlying distribution of
latent sources.

\begin{figure}
\begin{center}
  \resizebox{7cm}{!}{\includegraphics{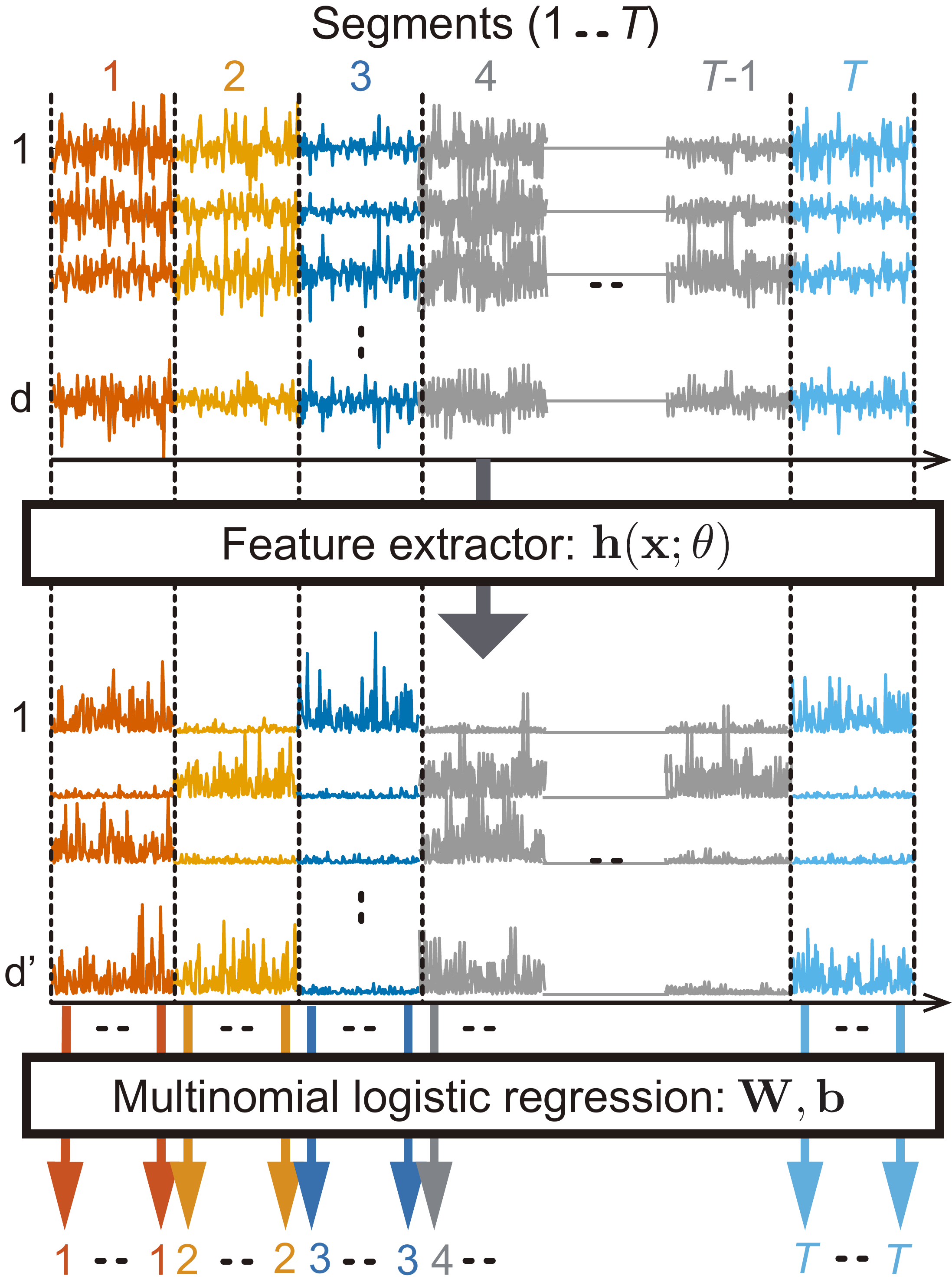}}
  \caption{Illustration of time-contrastive learning. The time points are segmented, as shown in different colors here. Each time point goes through a feature extractor $\hb$ which feeds the features to a multinomial regression layer. Together they learn to tell for each data point which segment it is from. Rather surprisingly, the feature extractor learns the independent components. } \label{tcl.fig}
\end{center}
\end{figure}

The theory of TCL~\citep{Hyva16NIPS} shows that the method can learn the independent
components up to
pointwise nonlinear transformations given by the $q$ above,
and a linear transformation $\mathbf{A}$. This is rather surprising since this SSL method does not make any reference to independent components.
A further linear ICA can recover the linear mixing $\mathbf{A}$ if
the number of segments grows to infinity and the segment distributions are random
in a certain sense. %
Thus, the theory proves that TCL (when supplemented by linear ICA) is consistent in the sense of estimation theory: when the number of data points grows infinite, the method finds the right components up to the point-wise nonlinearities. Such a consistency proof of the algorithm also implies identifiability of the underlying model, since for an unidentifiable model, a consistent estimating algorithm cannot exist. One caveat is that such statistical theory assumes that the optimization does not fail by getting stuck in a local optimum, which is, however, a typical practical problem in deep learning.
This SSL scheme is illustrated in Fig.~\ref{tcl.fig}.

Furthermore, with this self-supervised approach it is possible to combine estimation of components with dimension reduction. Heuristically, one can simply have a smaller dimension $d'$ in the feature extractor than the dimension of the data. This can be given a rigorous probabilistic interpretation, if we assume some of the components are actually uninteresting, ``noise'' components, characterized by being stationary, unlike the actual components~\citep{Hyva16NIPS}. The feature extractor will then simply ignore those dimensions which are not nonstationary.

In the basic model, it is assumed that the segmentation is known, or manually imposed. This may be a restriction in practice, although the method seems to work even if the segmentation is not very well specified. However, it is also possible to model the segmentation by a Hidden Markov Model~\citep{Halva20UAI,Halva21}, and estimate both the segmentation and the demixing by maximum likelihood, which will be considered below.

\subsection*{Permutation-Contrastive Learning}
Another approach to nonlinear ICA is to use the temporal dependencies of the independent components. Using the (linear) autocorrelations of stationary sources enables separation of the sources in the linear mixing case, although under some restrictive conditions~\citep{Tong91,Belo97}. A major advance in the field was to show how this framework can be extended to the nonlinear case~\citep{Harmeling03,Sprekeler14}. Related proposals have also been made under the heading ``slow feature analysis''~\citep{Wiskott02,Foldiak91}. Recent deep learning research~\citep{mobahi2009deep,springenberg2012learning,goroshin2015unsupervised} %
uses similar ideas, often called ``temporal coherence'', ``temporal stability'', or ``slowness'' of the features. Lack of rigorous theory has been a major impediment for development of such methods in the nonlinear case. 

The framework of Permutation-Contrastive Learning (PCL)~\citep{Hyva17AISTATS} enables a rigorous treatment of the identifiability of such models in a nonlinear generative model setting. 
The basic idea is to assume that the independent components are again time series, but this time they are \textit{stationary} and have \textit{temporal dependencies}. As a simple example of great practical utility, each independent component might follow a possibly nonlinear autoregressive process with possibly non-Gaussian innovations, given in the basic case of one time lag by
\begin{equation} \label{AR}
  s_i(t)=r_i(s_i(t-1))+n_i(t)
\end{equation}
for some scalar autoregressive function $r_i$, and an innovation process $n_i(t)$. 

An intuitive justification for why such a model is identifiable is that the model imposes \textit{independence over all time lags}, i.e., between $s_i(t)$ and $s_j(t-\tau)$ for any $\tau$. This is another way of creating more constraints for finding the independent components~\citep{schell2023nonlinear}. Thus, it is intuitively plausible that we get rid of the non-uniqueness of the basic i.i.d.\ case. The proof assumes certain conditions which essentially mean that the components are sufficiently temporally dependent and non-Gaussian~\citep{Hyva17AISTATS,Halva21,schell2023nonlinear}. For example, in the autoregressive model in Eq.~(\ref{AR}), as soon as either $r_i$ is nonlinear, or $n_i$ is non-Gaussian, the model is identifiable~\citep{Hyva17AISTATS}. (If the autoregressive model is linear and Gaussian, separation is possible but only if the autocorrelations are different from one component to another~\citep{Tong91,Belo97,Sprekeler14}.)

\begin{figure}
\begin{center}
  \resizebox{7cm}{!}{\includegraphics{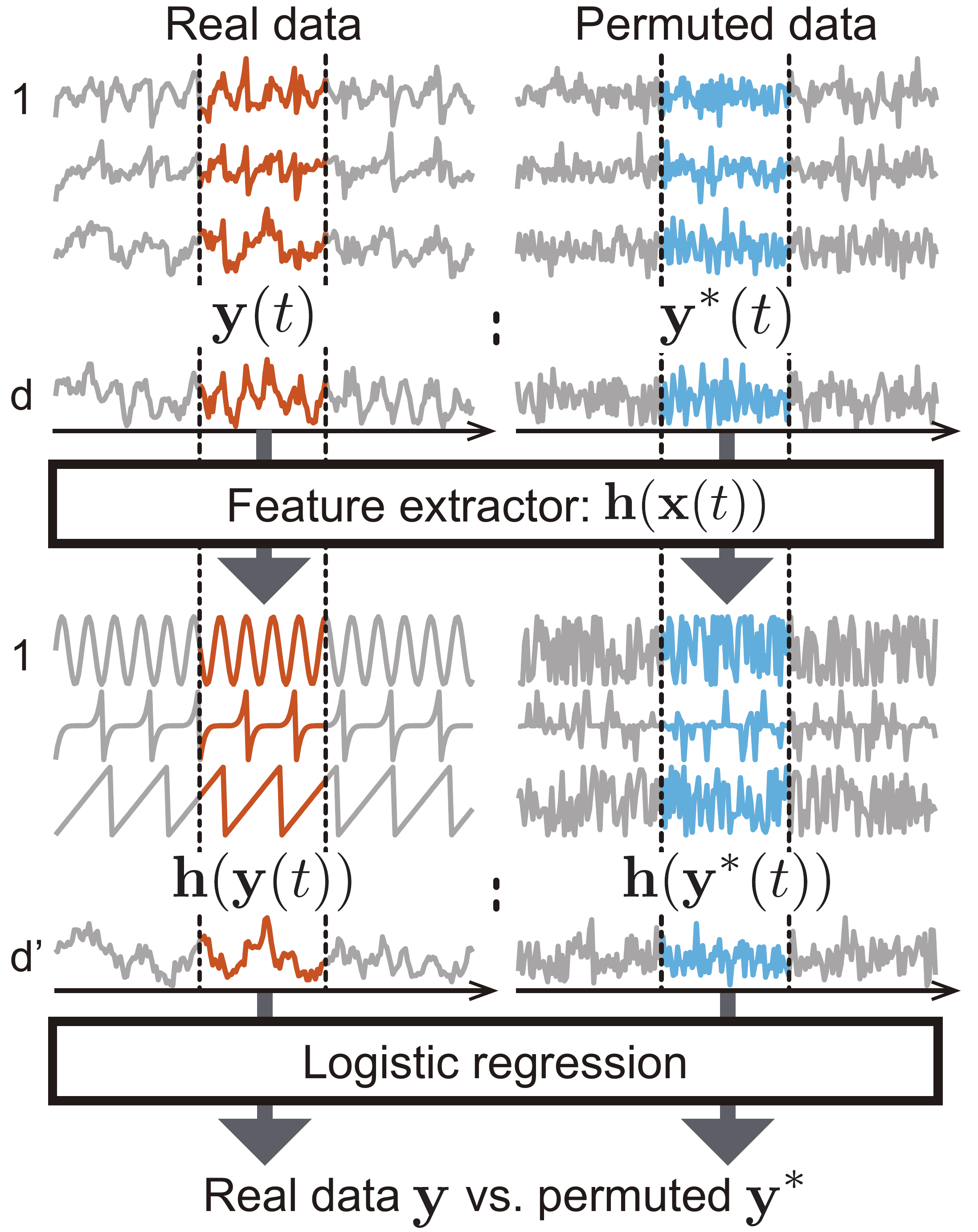}}
  \caption{Illustration of permutation-contrastive learning. Short windows $\y$ are sampled from the data (one window is given in red). Likewise, windows are sampled from data where the time dependencies are destroyed by random time-permutation. The feature extractor, together with a binary logistic regression classifier, learn to tell which input $\y$ is real and which is permuted (randomized). In the process, the feature extractor learns the independent components. }\label{pcl.fig}
\end{center}
\end{figure}

PCL is also an SSL algorithm, which in the simplest case proceeds as follows. Collect data points in two subsequent time points to construct a sample of a new random vector $\y$:
\begin{equation} \label{y}
\y(t)=\begin{pmatrix} \x(t) \\ \x(t-1) \end{pmatrix}
\end{equation}
which gives a ``minimal description'' of the temporal dependencies in the data.  Here, the same $t$ is used as the sample index for $\y(t)$ as for $\x(t)$. 
As a contrast, create a \textit{permuted} data sample by randomly permuting (shuffling) the time indices:
\begin{equation} \label{ystar}
\y^*(t)=\begin{pmatrix} \x(t) \\ \x(t^*) \end{pmatrix}
\end{equation}
where $t^*$ is a randomly selected time point. In other words, we create data with the same marginal distribution (on the level of the vectors $\x$ instead of single variables), but which does not reflect the temporal structure of the data at all.
Next, learn to discriminate between real data $\y(t)$ and time-permuted data $\y^*(t)$. 
We use logistic regression with a regression function of a special form, where like in TCL, the neural network can be divided into a feature extractor $\mathbf{h}$ and the final logistic regression layer.
The learning system is illustrated in Fig.~\ref{pcl.fig}.
We note that very similar ideas have been proposed heuristically in more applied contexts~\citep{misra2016shuffle,Banville21}.

Intuitively speaking, it is plausible that the feature extractor $\hb$ somehow recovers the temporal structure of the data since recovering such structure is necessary to discriminate real data from permuted data. In particular, since the most parsimonious description of the temporal structure can be found by separating the sources and then modelling the temporal structure of each source separately, it is plausible that the discrimination works best when the $h_i$ separate the sources. Like with TCL, the theory of PCL\citep{Hyva17AISTATS} rigorously proves that the algorithm is actually statistically consistent, and that dimension reduction is possible by just using a smaller number of hidden units.

In Fig.~\ref{separation.fig}, we show that PCL performs nonlinear ICA: The bottom row shows the results of PCL applied on the nonlinear mixtures of the four original signals. Clearly, PCL found very good approximations of the original signals. In fact, these four signals are characterized by temporal dependencies, and they are not Gaussian processes, so the conditions of PCL can be fullfilled. Likewise, Fig.~\ref{ident.fig} d) shows the results of applying PCL on the data in Fig.~\ref{ident.fig} a); the data was actually generated with temporally dependent component although that could not be seen in the figure. Again, PCL finds the original components up to small errors.

\subsection*{Combining temporal dependencies and nonstationarity}

While TCL and PCL probe two different kinds of temporal structure, nonstationarity and temporal dependencies respectively, it would be of great interest to develop a model and an algorithm that combines the two.
It is in fact rather straightforward to combine these two properties in a single statistical model~\citep{Halva21}. In the simplest case, we simply augment the AR model in Eq.~(\ref{AR}) so that the variance of the \textit{innovation} is nonstationary:
\begin{equation} \label{ARnonstat}
  s_i(t)=r(s_i(t-1))+\sigma_i(t)n_i(t)
\end{equation}
for some nonstationary signal $\sigma_i(t)$ which follows another AR (or Markov) model, thus constituting a Hidden Markov Model (HMM)~\citep{Halva20UAI,Halva21}. This basic idea has been generalized to create a very general framework called Structured Nonlinear ICA, or SNICA~\citep{Halva21}. Maximum likelihood estimation can be used to learn such a model, as will be explained below.

An SSL framework that incorporates the same idea is Independent Innovation Analysis (IIA)~\citep{Morioka21AISTATS}. In fact, it goes a bit further and proposes a model in which it is not the components themselves but the innovations that are independent. The method starts by assuming a completely general autoregressive model as
\begin{equation}
  \x(t)=\f(\x(t-1),\s(t))
\end{equation}
where the $\s(t)$ now take the role of nonlinear innovations, and $\f$ combines the autoregressive function and the mixing function. Like in basic nonlinear ICA, we aim to estimate $\f$ and $\s$. The entries of $s_i(t)$ are assumed independent, and typically nonstationary, although different assumptions are possible here. An SSL method can then be develop by considering the augmented data vector $(\x(t),\x(t-1))$ and applying a variant of TCL on it.

Contrastive Predictive Coding (CPC)~\citep{oord2018representation} is a related self-supervised method. The system learns nonlinear features from time windows, and models the distributions of the latent components based on those features as well as a latent ``context'' variables  which summarize the history of the time series. The context model is not unlike the nonstationarity model with a HMM. However, we are not aware of work directly connecting CPC to a identifiable latent-variable model.

\subsection*{Nonlinear ICA using auxiliary variables}
Another generalization~\citep{Hyva19AISTATS} of nonlinear ICA
is to assume that each component $s_i$ is dependent on some \textit{observed}
\textit{auxiliary variable} $\ub$, but independent of all the other
components, conditionally on $\ub$:
\begin{align*}
    p(\s\vert\ub) &= \prod_i p_i(s_i\vert\ub).
\end{align*}
which is to be compared with the basic independence in Eq.~(\ref{factorize}).
This formulation is so general that it subsumes TCL  as a special case:
in the case of nonstationary sources, the auxiliary variable $\ub$
can be the segment label.
More generally, the auxiliary variable $\mathbf{u}$ can be a class label, the
index of a pixel in an image, some description of an image, the sound of a
video~\citep{arandjelovic2017look} amongst others. Related approaches assume we have multiple views of the same data~\citep{gresele2020incomplete} or the data is multimodal~\citep{Morioka23AISTATS}.
A clear connection to PCL can be made as well, by considering $\ub$ to contain the history of $\x(t)$, perhaps simply $\x(t-1)$. Thus, we see that nonlinear ICA is possible without any time structure, at the expense of having some additional observed data in the form of $\ub$.

Various estimation methods have been developed for this model.
A method called Generalized Contrastive Learning~\citep{Hyva19AISTATS}
learns the demixing function using a self-supervised binary discrimination task based on randomization reminiscent of PCL:
new data is constructed from the observations $\xb$ and $\ub$ to obtain two data sets
\begin{align*}
	\tilde{\xb} &= (\xb, \ub), \\
	\tilde{\xb}^* &= (\xb, \ub^*),
\end{align*}
where $\ub^*$ is drawn randomly from the distribution of $\ub$ and independent of $\xb$.
Then, nonlinear logistic regression is performed using a regression function of a specific form
to discriminate between actual samples $\tilde{\xb}$ and shuffled samples $\tilde{\xb}^*$. 
The intuitive justification is that according to the generative model,
the observed and the auxiliary variables in the non-shuffled data set $\tilde{\mathbf{x}}$
are linked through shared latent variables,
whereas this link is broken in the shuffled data set $\tilde{\mathbf{x}}^*$.
Thus, the regression function makes use of
a feature extractor $\hb$
like in TCL and PCL, the purpose of which
is to extract the latent features
that allow distinguishing between the two data sets.
The theory~\citep{Hyva19AISTATS,Khemakhem20iVAE} shows that
the model is identifiable up to component-wise invertible transformations, and that the estimator given by this self-supervised method is consistent, 
provided that the latent distribution $p_i(s_i\vert\mathbf{u})$ satisfies some
regularity constraints.

\subsection*{Maximum likelihood estimation}

Once a probabilistic model for nonlinear ICA has been defined, it should be possible to estimate it by maximization of likelihood. This provides an alternative to SSL for estimation of each of the probabilistic models discussed above.
Maximum likelihood estimation is statistically optimal in the sense that it is asymptotically efficient (achieves the smallest statistical error for a finite data set) under mild conditions. Another the benefit of maximum likelihood methods is that they can be seamlessly integrated with further probabilistic inference. For example, the actual values of components can be inferred in case of observational noise\citep{Halva21}, or a segmentation of the time series can be inferred simultaneously with the estimation\citep{Halva20UAI}. (But it is also possible to first estimate the mixing function by SSL, and do probabilistic inference of latent quantities afterwards.)

The problem with maximum likelihood estimation is that it can be computationally very demanding. This is in contrast to the self-supervised methods presented above which tend to be algorithmically simpler while statistically less optimal. It is an empirical question which class of methods is better for nonlinear ICA; a general answer can hardly be given since it depends on the data set being analyzed as well as the computational environment being used. 

Maximum likelihood methods have been developed for nonlinear ICA based on two different approaches. The first  possibility is to consider the (exact) likelihood of the models considered above. In the case of time-series, the likelihood of such models can be generically expressed in a simple formula:
\begin{equation}
  \log p(\x(1),\ldots,\x(t_\text{max});\g)=\sum_{i=1}^d \log p_i(g_i(\x(1)),\ldots,g_i(\x(t_\text{max})))+\sum_{t=1}^{t_\text{max}} \log|\det \J\g(\x(t))|
\end{equation}
where the important point is the apparition of the determinant of the Jacobian $\J\g$ of the demixing function. The first term on the right-hand-side presents no particular difficulties regarding its computation and optimization: it  is simply the likelihood given by the time series model (e.g., autoregressive) for each estimated component, for the whole time series with time index from $1$ to $t_\text{max}$. But the apparently simple determinant in the second term is computationally very difficult to optimize when $\g$ is a neural network. This problem has been extensively considered in the case of normalizing flows~\citep{kobyzev2020normalizing}, where the typical solution is to strongly constrain the function $\g$, for example so that it has a triangular Jacobian. However, in our case we don't want to constrain the function $\g$ in any way because we want to be able to estimate general nonlinear mixing functions. Thus, the solutions offered by the literature on normalizing flows is of little use here.

Fortunately, it is possible to use what is called the (Riemannian) relative gradient for computationally efficient optimization~\citep{Gresele20}. Thus, maximum likelihood estimation of the noise-free model becomes  possible in practice. This would seem to be the statistically ultimate method in the sense of being not only consistent, but even asymptotically efficient. The downside is that this method requires the number of independent components to be equal to the number of observed variables, not allowing for simultaneous dimension reduction, unlike almost all the other methods considered in this paper. In practice, the dimension would need to be first reduced by some other method similar to PCA, as is almost always done with linear ICA.

Another possibility is to use variational approximations of the likelihood. This assumes we add a noise term to the mixing, as typical of VAEs in Eq.~(\ref{VAE}), since otherwise the posterior distributions are degenerate and variational methods do not work. Such variational methods are basically variants of the well-known VAE framework. Thus, this approach also shows how a VAE can be made identifiable: either by looking at time structure, which results in the SNICA~\citep{Halva21} and SlowVAE~\citep{klindt2020towards} models, or by conditioning by an auxiliary variable, which results in a very general model called iVAE\cite{Khemakhem20iVAE}. In both cases, VAE gives a variational method for estimation of the model.

Such variational methods have the advantage that they can reduce the dimension of the data at the same time as estimating components; above we argued that dimension reduction is actually the main utility of the plain VAE. This is an improvement on the noise-free maximum likelihood considered above, but the self-supervised methods are also able to reduce the dimension. On the negative side, variational methods are based on approximations and thus unlikely to be statistically consistent (i.e., to converge to the right solution in the limit of infinite data). Depending on the data, the bias introduced may be negligible with respect to the gain in statistical efficiency compared to self-supervised methods, or it may not be so.

Another related approach is \textit{energy-based} modelling~\citep{song2021train}. It can also be used for probabilistically principled estimation of the nonlinear ICA model~\citep{Khemakhem20NIPS}. Instead of the likelihood, some other objective is maximized (e.g., score matching distance), and the latent variables are not explicitly given by the model. The estimation is thus greatly simplified at the expense of losing some of the benefits of maximum likelihood estimation. This may offer an interesting compromise between statistical efficiency and computational efficiency. Finally, let us mention that the principle of adversarial learning as in GANs can also be used to estimate the nonlinear ICA model~\citep{LuopajarviThesis}: just like VAE, GAN can in fact be seen as a general principle for estimating a latent-variable model.

\section*{Discussion}

In this paper, we reviewed recent research on nonlinear ICA. It provides identifiable models, i.e., probabilistic models for which a unique solution can be shown to exist (up to trivial indeterminacies), and this solution finds the original components postulated in the model. While in many other fields of machine learning, identifiability is not a problem, it is a fundamental problem in the case of finding hidden factors, or disentanglement, as has been known since the 1950s at least. Identifiability can be attained for nonlinear ICA by postulating time series, or observing an additional auxiliary variable. After defining an identifiable model, different algorithms can be devised; we focused here on self-supervised algorithms and maximum likelihood estimation (including variational methods).

\subsection*{Applications}

\begin{figure}
\begin{center}
  \resizebox{\textwidth}{!}{\includegraphics{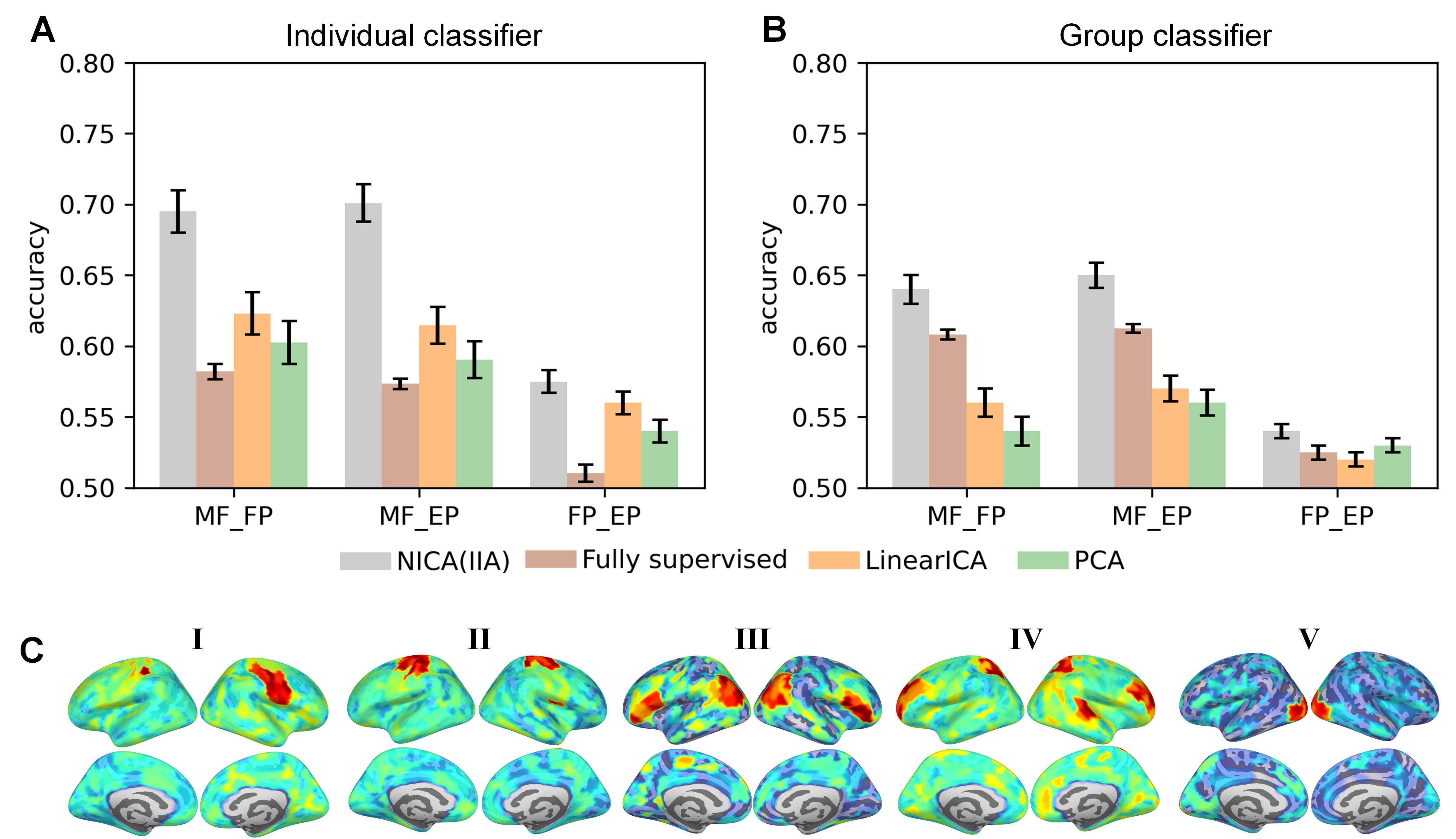}}
  \caption{Application of nonlinear ICA to brain imaging data in a semi-supervised setting\citep{Zhu23}. Magnetoencephalographic (MEG) data consisted of 306 measured time signals depicting brain activity. Nonlinear ICA was applied using IIA on a publicly available unlabelled, general-purpose, big data set with measurements from 652 participants. A-B) Given a smaller, specialized data set with 24 participants engaged in a mindfulness meditation task (MF) and two control tasks (FP, EP), a linear classifier was trained to ``decode'' (classify) the task from the MEG data.
    Using the features learned by nonlinear ICA (NICA) substantially increased the ability to decode if the participant was engaged in meditation (MF vs.\ FP and MF vs.\ EP),  compared to both a fully supervised classifier, and linear feature baselines (ICA, PCA). The classifier was trained either for each participant separately (A), which gave much better results, or for all the participants together (B). 
    C) A basic visualization of the spatial patterns of five selected nonlinear independent components; however, it should be noted that the workings of a neural network is notoriously difficult to interpret~\citep{Zhang2021survey}. Overall, these results indicate that building a neurofeedback device to help in mindfulness meditation~\citep{Zhigalov19} would benefit from semi-supervised learning, in particular feature extraction by nonlinear ICA.
    }\label{zhufig}
\end{center}
\end{figure}

Since nonlinear ICA is a very recent method, its utility for real data analysis is still largely to be explored. Analysis of EEG and MEG data has already shown that nonlinear ICA~\citep{Morioka21AISTATS,Zhu23} or very closely related self-supervised methods~\citep{Banville21} provide a representation which is very useful for classification. This work typically considers a \textit{semi-supervised} setting\citep{Khemakhem20NIPS,Zhu23}, which means that a representation is learned from large unlabelled data sets in a unsupervised manner, and the learned neural network is then applied on a new labelled data set to compute features which are useful for classification. The point is that it is often easy to find a big data set which is unlabelled, while the data sets where the classification is of practical  significance are often small. Especially when learning a representation with a deep neural network, it is crucial to be able to do it from a big data set. This is closely related to \textit{transfer learning}, where the learned features are used on a \textit{new} data set\citep{Khemakhem20NIPS}. In neuroscience, another utility of such features is that they can provide insight into the structure of the data~\citep{zhou2020learning,schneider2023learnable}, with the caveat that interpretation of neural networks is notoriously difficult. An application to brain imaging data is shown in Fig.~\ref{zhufig}.

Learning image features is an application that holds great promise~\citep{klindt2020towards}. This should be straightforward using video as input data, but requires huge computational resources. In fact, a large number of SSL methods have been proposed in the context of computer vision, and some of them are almost identical to nonlinear ICA~\citep{misra2016shuffle,arandjelovic2017look}. Thus, the theory of nonlinear ICA could also be seen as a post hoc theoretical justification of some of the existing self-supervised methods in computer vision. Such self-supervised methods are often completely heuristic, and building a proper theory to explain their behavior is most interesting.  Audio data has great similarities to image and video data, so nonlinear ICA  is likely to work there as well, even if existing methods tend to use audio- or speech-specific methods~\citep{ravanelli2020multi}.

An interesting question is whether nonlinear ICA could provide a successful method for generating new data points. It should be particularly suitable for conditional generation, since many models assume some kind of conditioner (auxiliary variable). The promise  would be that the independent components correspond to some meaningful quantities, so that it would be useful to explicitly manipulate them in some use cases. To our knowledge, this has not been seriously investigated at the moment.

Another extremely interesting application can be found in causal discovery. Nonlinear ICA allows for the determination of the direction of effect, i.e., which variable causes which, even in completely nonlinear regression models~\citep{Monti19UAI}. Here, identifiability is absolutely essential since the very point of the analysis is to interpret the parameters estimated, as they express the direction of effect. For details, we refer to our review on identifiability theory~\citep{Hyva23AISM}.

It should be emphasized, however, that due to the great generality of the nonlinear ICA model, it can be applied in many different domains. The most successful applications may be something quite different from what was just discussed.

\subsection*{Extensions}

Here we focused on models that achieve identifiability by statistical assumptions: assuming temporal or other dependencies inside the components. An alternative is to consider restrictions on the nonlinearity~\citep{gresele2021independent,zimmermann2021contrastive,buchholz2022function,kivva2022identifiability,moran2021identifiable}. In earlier research, some very strong restrictions have been proposed, in particular constraining the mixing to be linear followed by point-wise nonlinearities~\citep{taleb1999source}. However, in the case of general, non-parametric nonlinearities, it is not quite clear what kind of conditions would be flexible enough to be useful, while enable identifiability~\citep{Hyva23AISM}.

Models for dependencies \textit{between} the components have also been developed~\citep{Khemakhem20NIPS}. A special case, and a topic of great current interest, is how to combine estimation of components with analysis of their causal relations, leading to causal representation learning~\citep{lachapelle2022disentanglement,Morioka23AISTATS}. Indeed, finding models which allow for both a general nonlinear (observational) mixing and some kind of (causal) dependencies between the components are of great interest in future research. At first sight, there may seem to be some inherent contradiction between finding components which are dependent, since any such dependencies might be modelled by the mixing. However, a possible resolution to this contradiction might be to use more than one of the statistical principles given above. For example, the components might be identifiable based on their temporal dependencies, while the causal relations might be found, perhaps afterwards, by looking at their nonstationary dependencies~\citep{Monti19UAI,Zhang10UAI}, or even instantaneous dependencies~\citep{Khemakhem20NIPS}; multimodal (three-way) data allows for another approach~\citep{Morioka23AISTATS}.

The very definition of identifiability can also be extended. In this paper, we used a definition typical in linear ICA research, where some indeterminacies (order of components, scaling, and signs) are allowed. Identifiability without any such indeterminacies can also be considered~\citep{xi2023indeterminacy}. On the other hand, our theory assumed infinite data as well as a universal function approximator, which are both unachievable in a real learning scenario. 
Estimation errors in such practical scenarios can be analyzed,  resulting in an upper bound which is a function of the complexity and the smoothness of the demixing function, in addition to sample size~\citep{lyu2022finite}. It is, in fact, intuitive that learning a more complex demixing requires more data, and that lack of smoothness makes the estimation harder as well.

\subsection*{Self-supervised learning vs.\ nonlinear ICA}

In this paper, nonlinear ICA was largely approached from the viewpoint of SSL. However, it is important to note that there is no logically necessary connection between the two. On the one hand, it is not at all necessary to use SSL for nonlinear ICA. In fact we discussed methods using maximum likelihood estimation above, and those methods are not self-supervised. On the other hand, a huge number of SSL methods have recently been proposed and only a tiny proportion is related to estimation of an ICA model, or of any probabilistic model at all. SSL methods can even be used for purposes completely unrelated to feature extraction, such as learning to approximate probability distributions~\citep{Gutmann12JMLR,vincent2011connection}. 

It is in fact important to draw the distinction between the probabilistic model and its estimation method. While nonlinear ICA is fundamentally based on defining a probabilistic model, the model also needs an algorithm for its estimation, and that can be provided by SSL (or maximum likelihood). However, an SSL algorithm never provides, in itself, a probabilistic model. Furthermore, identifiability is always a property of the probabilistic model, not an estimation algorithm. This distinction is slightly confounded in the context of methods such as TCL and PCL where a probabilistic model is introduced in the same paper as an SSL estimation method. The identifiability proofs may even rely on proving the convergence of the algorithm, but this is just one proof technique that can be replaced by proofs making no reference to any estimation method~\citep{Khemakhem20iVAE,Halva21}. 

The fact that SSL methods have been important in the field of nonlinear ICA can be understood from the viewpoint that SSL is a very flexible and computationally appealing framework for performing unsupervised learning, and perhaps mainly by chance, most of the initial work on nonlinear ICA used SSL. A more recent thrust has been to use purely probabilistic methods and maximum likelihood as reviewed above. It remains to be seen which approach will be more successful in practical applications.

\subsection*{Goals of nonlinear ICA and unsupervised learning}

While we might casually just say that nonlinear ICA is a method for unsupervised learning, we think it is important to note that unsupervised learning can have different goals. While different lists of goals have been given~\citep{theis2015note}, we propose to consider the following four:
(Goal 1) Estimating an  accurate model of the data distribution. Energy-based modelling in terms of score matching or noise-contrastive estimation, as well as VAEs and normalizing flows are fundamentally designed for this purpose. (Goal 2) Sampling points from the data distribution. Generative adversarial networks (GANs) were conceived for this very purpose, although more recently, generative diffusion models may have been more successful. (Goal 3) Obtaining useful features or a representation for supervised learning. Here, we come to the question of representation learning which was one of the starting points of this paper. While it is often performed by VAEs and even GANs, the problems of identifiability reviewed here suggest that nonlinear ICA should be better. (Goal 4) Revealing underlying structure in data. In this case, the question of identifiability becomes paramount: The features learned cannot be meaningfully considered to reveal the underlying structure unless the model is identifiable. This is particularly important in scientific data analysis where the features are often assumed to correspond to some scientifically interesting quantities.

Importantly, these four goals are partly orthogonal, even contradictory. In particular, goals 1 and 2 are essentially non-parametric problems which require an arbitrarily good approximation of the probability distribution. Such an approximation can be very well done by a black box whose inner working are neither understood nor computationally accessible. In contrast, for goals 3 and 4, we need a system which is not a black box, and a model parametrized by a judiciously chosen, possibly low-dimensional parameter vector may be best.  Therefore, it seems unlikely that any single method could accomplish all of these goals. We propose that in unsupervised learning research, one should specify the more specific goal; unsupervised learning in itself is not a properly defined goal. As for nonlinear ICA, the primary goals are 3 and 4: learning a good representation for supervised learning and revealing the underlying structure of multi-dimensional data.

\subsection*{Resource availability}

Software implementations are available for several of the methods described in this paper, \\see {\tt https://www.cs.helsinki.fi/u/ahyvarin/software.shtml}.

\section*{Acknowledgments}

A.H.\ was supported by a CIFAR Fellowship and the Academy of Finland.
H.M.\ was supported in part by JST PRESTO JPMJPR2028, JSPS KAKENHI 22H05666 and 22K17956.

\section*{Declaration of interests}

I.K.\ is currently employed by Marshall Wace LLP, but his contribution to this article is entirely based on his earlier PhD work at UCL prior to joining Marshall Wace. Marshall Wace had no role or influence in writing this article.

\section*{Author Contributions}

Conceptualization, A.H.;
Writing -- Original Draft, A.H., I.K.;
Writing -- Review \& Editing, A.H., H.M.;
Visualization, A.H., H.M.;
Supervision, A.H.


\end{document}